\documentclass{article}
\usepackage[utf8]{inputenc}
\usepackage{graphicx}
\usepackage{amsmath, amssymb}
\usepackage{cite}
\usepackage{hyperref}
\usepackage{booktabs}
\usepackage[a4paper, total={6.5in, 9in}]{geometry}
\usepackage[most]{tcolorbox}
\usepackage{xurl}
\usepackage{fontawesome5}
\usepackage{paracol}
\tcbuselibrary{breakable, skins}
\newtcolorbox{mainbox}{
  enhanced,
  colback=blue!2,
  colframe=blue!35,
  boxrule=0.8pt,
  arc=4mm,
  left=7mm,right=7mm,top=7mm,bottom=7mm
}
\newtcolorbox{sidebox}[3]{
  enhanced,
  colback=#1!5,
  colframe=#1!65,
  boxrule=0.8pt,
  arc=2mm,
  left=3mm,right=3mm,top=2mm,bottom=2mm,
  title={#2\quad #3},
  fonttitle=\bfseries\Large,
  coltitle=white,
  colbacktitle=#1!70
}

\title{\textbf{TeleMorpher: Toward Robust Simultaneous Motion-Location Editing}}

\author{
{\large Haengbok Chung}\\
[0.3em]
}
\date{}

\begin{document}

\maketitle

\begin{tcolorbox}[
    enhanced,
    colback=white,
    colframe=blue!35,
    arc=3mm,
    boxrule=0.8pt,
    width=\textwidth,
    left=6mm,right=6mm,top=5mm,bottom=5mm
]

\begin{center}
{\Large\bfseries Abstract}
\end{center}


\vspace{2mm}

\noindent
Diffusion models have achieved remarkable success in image and video generation and editing. While recent studies have extended these efforts toward motion editing, simultaneously transforming both motion and location—despite its practical importance—remains largely unexplored. To better understand robust motion-location editing, we first analyze the fundamental factors that degrade its quality. Based on this analysis, we propose \textbf{TeleMorpher}, one of the first one-shot frameworks to the best of our knowledge, for simultaneous motion-location editing. Our approach leverages motion priors, a target motion-centric video generated from an off-the-shelf model as motion-editing guidance, and the ground truth motion to enable more controllable and precise motion-location editing. Via this, our framework works as follows: (1) we first disentangle the protagonist and the background via pre-trained segmentation and inpainting models. (2) Then, we introduce a training-free pose warping that edits the protagonist's motion with the motion prior as the guidance. (3) The result of warped motion video is directly injected into a baseline motion editor during inference, mitigating the difference between source and target motions while preserving the appearance of the source video. (4) To enhance the reliability of quantitative evaluations, we propose two new LPIPS-based metrics that measure the background consistency before and after the motion editing and the fidelity of motion editing performance via measuring the difference between the extracted protagonist's skeletons from source and target videos. Experiments with in-the-wild videos and the TaiChi dataset demonstrate that TeleMorpher achieves superior performance across both quantitative and qualitative measurements (real-human evaluation), underscoring its effectiveness.

\vspace{5mm}

\noindent

\begin{minipage}[t]{0.96\textwidth}
\begin{tcolorbox}[
    colback=blue!5,
    colframe=blue!60,
    coltitle=white,
    colbacktitle=blue!60,
    arc=2mm]
\small
\href{https://github.com/Happiness-Chung/TeleMorpher.git}{\texttt{Github: https://github.com/Happiness-Chung/TeleMorpher.git}}
\end{tcolorbox}
\end{minipage}
\end{tcolorbox}

\section{Introduction}

\;\;\;\; Diffusion-based generative models have emerged as a powerful approach for producing high-quality videos \cite{1,2,3,4,5}. With the foundation of success in video generation, video editing techniques for generated content have progressed at an accelerated pace \cite{6,7,8,9}. Among them, motion editing \cite{60, 61, 62, 63} emerged as one of the promising areas because of its potential for broad applications in video content creation, virtual character animation, gaming, robotics, sports analysis, healthcare, education, and data augmentation. 

Although significant progress has been made in efficient motion editing \cite{10,11}, no prior work has explicitly addressed simultaneous few-shot motion-location editing to the best of our knowledge. This is primarily due to the inherent challenges of the task, such as reducing the large difference between the source and target motions, preserving the background and the protagonist's appearance during the editing process, and maintaining temporal consistency across frames. Notably, to solve these problems, MotionEditor \cite{10} proposed a one-shot motion editing pipeline that consists of a content-aware motion adapter, integrated with ControlNet and an attention injection mechanism. This injects features of the source video into the neural network during the inference to better preserve the appearance of the background and protagonist in the source video. Edit-Your-Motion \cite{11} shuffles frames to mitigate the discrepancy between source and target motion conditions and adapts attention injection similar to MotionEditor. While these methods successfully pioneered motion editing with single-shot training for the first time, they often suffer from flickering, appearance inconsistency, and sub-optimal faithful editing performance. Moreover, they do not address location changes, which are essential for more practical protagonist editing in video. However, many real-world applications require simultaneous control of both motion and location.

To address these limitations and enable simultaneous motion-location editing, we explore the fundamental challenges that degrade its quality. Based on our findings, we propose a one-shot simultaneous motion-location editing pipeline, TeleMorpher. This pipeline suggests the usage of the motion prior as a ground truth, that is the synthetic motion video generated from the highly precisely controllable 3D avatars. This approach transfers the capability of the most precisely controllable 3D avatars \cite{45, 65} to motion-included videos recorded in the real or virtual world, without additional rendering or avatar controls. In addition, this pipeline proposes a new training-free motion-location-guidance generation method that boosts editing fidelity while preserving the appearances of video, and two new metrics that measure background preservation and skeletal motion alignment between edited and target motions. Via experiments using the 20 YouTube and synthetic \cite{50, 51} videos, and the TaiChi dataset \cite{64}, our framework demonstrates better controllability and more efficient motion-location editing than other baselines.

Our framework's contributions are summarized as follows:

\begin{itemize}
    \item We identify several fundamental challenges that degrade simultaneous motion-location editing via exploring diverse types of videos that contain humans.
    \item We introduce TeleMorpher, a new pipeline that leverages motion prior, a synthetic motion video generated from pre-trained models, for practical and easily controllable motion-location editing. 
    \item We propose a new training-free protagonist guidance generation method, that help reduce the gap between source and target motion while better preserving the appearance of the protagonist.
    \item Our experiments with 20 from the in-the-wild and synthetic videos and the TaiChi dataset demonstrate the promising potential of generalizable, robust, and high-quality motion-location editing across various domains. 
    \item In addition to the TeleMorpher code, to facilitate higher-quality motion-location editing, we publicly provide the collected videos and the results obtained from other baseline models as a TeleMorpher-smallpack. 
\end{itemize}

\section{Related Work}

This section reviews existing diffusion-based video editing models and provides an overview of pose and motion editing techniques across both image and video domains.

\subsection{Video editing with diffusion models}

Recent advances in diffusion models have extended their applicability image generation to video editing. To support effective editing across both spatial and temporal dimensions, existing methods can be categorized into several key approaches \cite{42}. Among them, attention adaptation methods \cite{12,13,14,15,29} aim to preserve temporal consistency by extending spatial diffusion models to temporal dimensions or by mixing keys and values from sparse frames. Structure-conditioning \cite{16,17,18,19} approaches incorporate intermediate structural cues, such as depth maps, bounding boxes, and appearance conditions, to guide generation and maintain spatial and temporal consistency. Meanwhile, training recipe methods \cite{20,21,22} modify training strategy by introducing motion-oriented loss functions or various fine-tuning methods. Attention feature injection techniques \cite{23,24,25} enhance motion controllability by capturing semantic correspondence into the generative process, often using cross-attention or self-attention. Diffusion latent manipulation methods \cite{26,27,28,30} include latent initialization, latent transition, underscoring smooth inter-frame transition. Despite these advances, precise and controllable editing—particularly for jointly modeling motion and spatial location—remains a challenging open problem.

\begin{figure*}
\begin{center}
\includegraphics[width=6.0in]{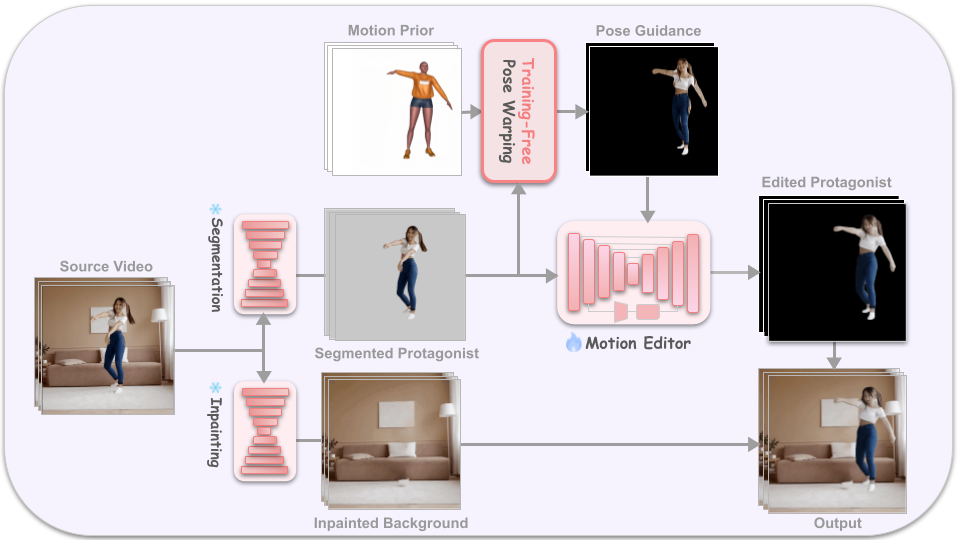}
\caption{Overview of the proposed pipeline (TeleMorpher). The input video is split into a protagonist and background video using a pre-trained segmentation model. The background is inpainted for later merging with the edited protagonist video. Motion-location editing is performed on the protagonist video using the motion priors instead of reference videos, enhancing controllability. A training-free pose warping step helps align motions better, preserving their appearance. Finally, the edited protagonist video is merged with the background to produce the final output.}
\label{fig:overall}
\end{center}
\end{figure*}

\subsection{Pose and motion editing with diffusion models}

Pose editing is a crucial task that enables fine-grained control for human movement, which is essential for realistic, controllable content generation in images and videos. Recent works \cite{31,32,33,34,35} have proposed methods for synthesizing new poses from a source human image and a condition, often by injecting appearance features from the source image to preserve identity consistency. Building on pose editing, motion editing enabled efficient, controllable video content editing. It aims to overcome key challenges such as temporal inconsistency, pose misalignment, and appearance drift \cite{42}. Early works \cite{36,37,38} introduced one-shot motion editing by leveraging pre-trained diffusion models trained on several static images. These methods often inject key-value pairs from the source video during inference to retain visual appearance while modifying motion. DiffBody \cite{60} integrates a 3D body prior into diffusion models, enabling substantial modifications to pose and body shape. TruePose \cite{61} employs human-parsing-guided attention mechanisms to better preserve facial identity and clothing details. DeCo \cite{62}, MotionFollower \cite{63} controls foreground and background in a decoupled manner for motion editing with the training procedure. However, these methods typically rely on a target video as a motion reference, suffering from temporal inconsistencies or appearance distortions when the target motion significantly deviates from the source video. In addition, training video diffusion models requires expensive resources. Although there is recent work in 3D virtual worlds that include highly controllable avatars, editing their motions requires additional rendering and manual controls, increasing the required resource overhead.

\begin{figure*}[hbt!]
\begin{center}
\includegraphics[width=\columnwidth]{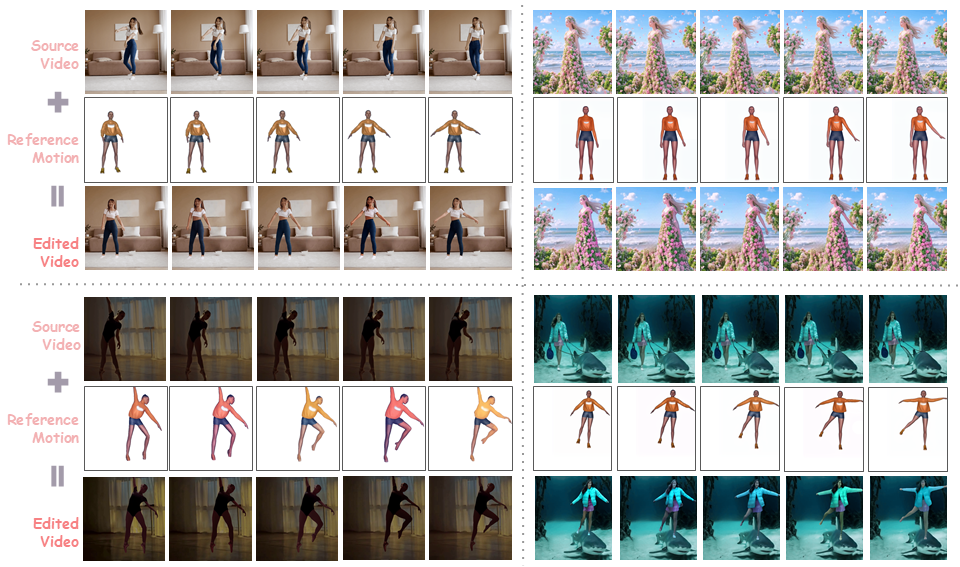}
\caption{We present qualitative examples showcasing how our framework successfully edits the protagonist’s motion while preserving appearance and seamlessly blending with the background. Each case demonstrates the ability to apply distinct target motions with high fidelity and realism.}
\label{fig:results}
\end{center}
\end{figure*}

\begin{figure*}
\begin{center}
\includegraphics[width=\columnwidth]{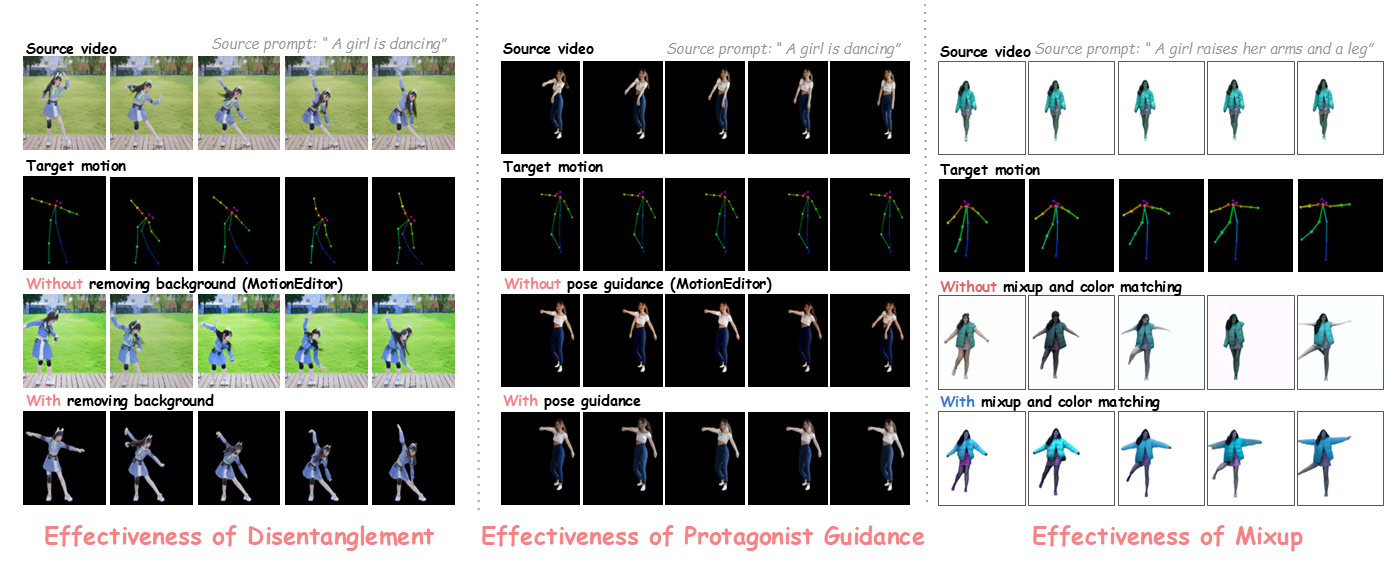}
\caption{Ablation of TeleMorpher to demonstrate the effectiveness of its components. From left to right: (1) background removal improves motion editing quality by reducing visual distractions, (2) pose-guided attention reduces flickering and enhances temporal coherence, and (3) mixing the protagonist condition with the input video further improves motion fidelity and consistency across frames.}
\label{fig:ablation}
\end{center}
\end{figure*}

\section{Preliminaries}

\;\;\;\; We provide an overview of video generation and editing methods guided by text and pose conditions. $c_{\text{text}}$ and $c_{\text{pose}}$ are the text and pose conditions. Given the conditions, a noise prediction network $\epsilon_{\theta}(\cdot)$ learns to predict video $v \in \mathbb{R}^{f X H X W X 3}$ corresponding to the given conditions, in which f is the number of frames, $H$ is the height, $W$ is the width of the video. To be specific, the model predicts a predefined noise $\epsilon \sim \mathcal{N}(0, I)$ with the diffusion time step $t \sim \mathcal{U}([1, ..., T])$. The objective function, therefore, minimizes the following loss:

\begin{equation}
    \mathbb{E}_{v,\epsilon, t}[|| \epsilon_{\theta}(v_t;c_{\text{text}}, c_{\text{pose}}, t) - \epsilon ||]
\end{equation}

After training, by replacing text and pose conditions with target conditions $\tilde{c}_{\text{text}}$ and $\tilde{c}_{\text{pose}}$, we can manipulate the pose of the protagonist by the following backward process. 

\begin{equation}
\begin{aligned}
p_\theta(v_{t-1} \mid v_t,\ \tilde{c}_{\text{text}},\ \tilde{c}_{\text{pose}}) \\
= \mathcal{N}(v_{t-1};\ \mu_\theta(v_t, t \mid \tilde{c}_{\text{text}}, \tilde{c}_{\text{pose}}), \sigma_t^2 \mathbf{I} )
\end{aligned}
\end{equation}
 
The mean \( \mu_\theta(v_t, t \mid \tilde{c}_{\text{text}}, \tilde{c}_{\text{pose}}) \) is predicted by a neural network 
conditioned on both modalities, while \( \sigma_t^2 \) is the predefined variance determined by the diffusion schedule.

Recent works, such as MotionEditor \cite{10} and Edit-Your-Motion \cite{11}, adopt one-shot strategies to effectively execute this editing process.

\section{TeleMorpher}
\subsection{Empirical Analysis of Motion-Location Editing Challenges}
Editing the protagonist's motion and location is a challenging task because of the degrading factors of the editing process. This section categorizes these factors into three main criteria: gap, ambiguity, and information amount, each with distinct subcategories that highlight the difficulties of the editing. These criteria provide a framework for systematic analysis of the relative difficulties inherent in the protagonist's motion-location editing task. Since it is inherently difficult to define absolute, objective thresholds for each factor, we roughly divide each factor's severity into two levels, expecting future refinements. This coarse but intuitive outlining enables extendable categorization of the challenges of this editing task, facilitating a clearer analysis. The experimental verification of this analysis is provided in the Appendix \ref{appendix:problem}.

\subsubsection{Gap}
The gap in motion editing arises primarily from the differences between the source and target videos. \textbf{Motion difference} refers to the extent of transformation from the source to target motions. At a basic level (Level 1), changes are limited to subparts of the body, such as a slight adjustment in hand or leg motion, which are relatively straightforward for algorithms to process. In contrast, at a more advanced level (Level 2), the entire body undergoes significant motion changes, requiring the system to capture and replicate challenging dynamics across frames. \textbf{Location} difference addresses the positional shifts of the protagonist within the frame. Maintaining visual coherence is less challenging when the protagonist's position remains consistent or undergoes minimal changes (Level 1). However, when the location shifts significantly beyond one-third of the image's length (Level 2), maintaining spatial continuity becomes a more intricate task, as the algorithm must manage spatial transformations while preserving other contextual details.

\subsubsection{Ambiguity}
Ambiguity in motion editing arises from the inherent difficulty in distinguishing precise details or boundaries in the visual data. One of the key factors is \textbf{resolution}, which affects the clarity of motion and object edges. Detailed information remains relatively distinct in cases of 512 x 512 resolution, which is relatively high (Level 1), enabling more accurate editing. However, when the resolution decreased to 256 x 256, which is relatively low (Level 2), it can obscure essential details, complicating the identification of motion patterns. Another critical factor is \textbf{human-background similarity}, which refers to how easily the protagonist can be separated from the background. When the colors and textures of the protagonist and background are distinct (Level 1), the diffusion model can easily process motion information, reducing potential errors. Conversely, when the colors or textures are similar (Level 2), the segmentation process becomes prone to artifacts and inaccuracies, requiring more sophisticated algorithms to achieve high-quality results.

\subsubsection{Information Amount}
The amount of information present in a video directly influences the complexity of motion editing. \textbf{Background complexity} is a key element, with simpler backgrounds (Level 1) featuring uniform color tones and minimal object diversity being easier to process. Challenging backgrounds (Level 2), with varying color tones and intricate object compositions, overwhelm the diffusion models and make it hard to focus on motion information. Another factor is \textbf{motion dynamicity}, which pertains to the rate of change in motion. Static or gradually changing motions (Level 1) are easier to manage as they require less precise temporal alignment. On the other hand, rapidly changing motions (Level 2) demand precise temporal synchronization and advanced predictive capabilities to ensure fluid transitions. Finally, \textbf{camera movement} introduces another layer of complexity. Maintaining alignment and coherence is relatively straightforward when the camera is fixed or exhibits minimal movement (Level 1). However, significant camera shifts (level 2) require compensatory adjustments in the editing process to align the motion and background seamlessly, increasing the technical demands on the editing system.

\subsection{Foreground and Background Disentangled Editing Framework}

While previous motion editing approaches \cite{10, 11} have introduced one-shot control of protagonist motion, they often focus on improving temporal consistency and reducing flickering rather than on what degrades quality. To further improve editing quality, we conduct a systematic analysis to find the root causes of performance degradation in motion editing and design a modular pipeline in which each component addresses a specific challenge. In particular, human-background similarity and background complexity both fundamentally arise from the presence of background content, which interferes with the editing process. To mitigate these factors, our pipeline performs motion editing without backgrounds, enabling the model to focus on the protagonist's motion in the source video. To enable this, as illustrated in Figure \ref{fig:overall}, TeleMorpher leverages pre-trained models to disentangle the background and the protagonist before editing.

By leveraging a pre-trained segmentation model, Segment Anything \cite{43}, we generate the protagonist mask $M_s$ for the source video $X_s$, then split the original video into the protagonist video $X_p$ and the background video $X_b$. Then, we fill the blank area in the background video for later use using the inpainting models. In this study, we used the Inpaint Anything \cite{44}.

Subsequently, we perform motion editing on the protagonist video. Unlike previous works that rely on existing real-reference videos, our approach utilizes synthetic motion priors $X_m$ \cite{45,46} as the target motion source. This makes the editing more precisely controllable and efficient, as users can flexibly define target motions without being constrained by several limitations: the shortage of existing video references that precisely contain motions that the user aims to transfer to the source video, additional rendering, or manual controls. In addition, compared to existing real-reference-video-based motion editing, these motion priors generated from controllable virtual avatar systems can provide substantially more varied and customizable motion reference options that are easier to obtain. As a result, motion priors significantly improve controllability, scalability, and editing efficiency while reducing the dependency on manually collected reference videos.

Then, to reduce the gap between the source and target motions while better preserving the appearance of the protagonist in the source video, we introduce a training-free pose warping strategy that generates a modified protagonist condition $P$, as explained in detail in Sections 4.3 and 4.4. The generated condition $P$ is fed to an existing motion editing baseline model to produce the edited protagonist video. Finally, we integrate the edited protagonist with the background video to generate the final output.

\subsection{Training-Free Protagonist Guidance}

Although our foreground–background disentangled motion-location editing pipeline improves its editing quality by eliminating background interference, pose misalignment between the source and target motions still remains a significant obstacle for accurate motion transformation. To address this challenge, we propose a training-free protagonist guidance method inspired by pose warping \cite{33} that enhances the conditioning effectiveness of the target motion. This strategy can be interpreted as a residual-style \cite{59} guidance mechanism, where the warped protagonist guidance fed to the network functions as a motion anchor for target motion during the diffusion process. Similar to how residual connections preserve input identity in deep networks, our injection mechanism helps maintain semantic consistency and appearance fidelity while boosting accurate motion transformation. While prior image-based pose editing works \cite{33} have adopted pose warping to guide structural transformation, these approaches typically require additional networks and training, resulting in increased computational overhead. To overcome this limitation and extend the idea to the video domain, we introduce a two-way motion warping strategy that operates without any additional training or network components. 

First, we prepare the source $M_s \in \{0,1\}^{F \times H \times W}$ and target segmentation mask $M_t \in \{0,1\}^{F \times H \times W}$ video using the off-the-shelf segmentation model. $F$ is the number of frames in the video, $H$ and $W$ are the height and width of the frames. Then we arbitrarily pick a frame $M_{p} \in \{0,1\}^{H \times W}$ from the source mask video. To identify and isolate regions of structural discrepancy, we compute the intersection of the source mask and target masks via element-wise multiplication. The overlapping area is retained as the shared region $M_{\mathrm{shared}}$, while the non-overlapping regions $M_{\mathrm{diff}}$-such as extended limbs or shifted body parts—are regarded as pose-specific differences. This procedure can be represented by Equation \ref{eqn:3} and Equation \ref{eqn:4}.

\begin{equation}
    M_{\text{shared}} = M_{p} \cdot M_t
    \label{eqn:3}
\end{equation}

\begin{equation}
    M_{\mathrm{diff}} = (M_p \cup M_t) \setminus (M_p \cap M_t)
    \label{eqn:4}
\end{equation}

In the Equation \ref{eqn:3}, $\cdot$ denotes element-wise multiplication. In Equation \ref{eqn:4}, Mdiff denotes the symmetric difference between $M_p$ and $M_t$, where $\cup$, $\cap$, and $\setminus$ denote the union, intersection, and set difference operators, respectively. To obtain $M_{\mathrm{diff}}$, we leverage the pose skeletons extracted from the source video and the target video with the target motion. It preserves regions that belong exclusively to either mask while removing their overlapping area, allowing us to identify pose-specific structural differences between the source and target motions.  Specifically, we retain only the areas in $M_{\mathrm{diff}}$ that semantically correspond to the target skeleton, such as arms and legs. This allows us to focus the transformation between corresponding body parts while discarding noisy or insignificant differences. Let $S_s \in \{0,1\}^{H \times W}$ and $S_t \in \{0,1\}^{F \times H \times W}$ denote the binary skeleton mask of the source and target motions. The refined motion-difference mask is defined as Equation \ref{eqn:5}, and Equation \ref{eqn:6}:

\begin{equation}
    M_{\text{pr}} = M_{\text{p}} \cdot S_s 
    \label{eqn:5}
\end{equation}

\begin{equation}
    M_{\text{tr}} = M_{\text{t}} \cdot S_t
    \label{eqn:6}
\end{equation}

Specifically, we identify pairs of body parts that indicate the same anatomical body part through the Equation \ref{eqn:5} and Equation \ref{eqn:6}, utilizing source and target masks of the protagonists (e.g., left arm, right leg). This allows us to obtain a set of $n$ sub-region-wise correspondences between the source and frames as well, denoted as $\{(M_{\text{pr}}^i, M_{\text{tr}}^i)\}_{i=1}^n$ for sub-region mask pairs and $\{(R_{\text{pr}}^i$, $R_{\text{tr}}^i)\}_{i=1}^n$ for sub-region image pairs where $n$ is the number of pairs. In these representations, each of $i^{th}$ sub-region pair corresponds to the same body part. 

For each corresponding pair of body-part $\{(M_{\text{pr}}^i, M_{\text{tr}}^i)\}_{i=1}^n$ we perform a warping from the source mask to the target mask. To this end, we propose two complementary warping methods that enable efficient training-free source to target transformation of each pair in $\{(R_{\text{pr}}^i$, $R_{\text{tr}}^i)\}_{i=1}^n$. 

The first warping method employs a parametric transformation by fitting a quadratic curve of the source sub-region mask to the target sub-region masks. For each corresponding pair, we extract the geometric centerline-based second-order polynomial from the source sub-region mask and fit it to the corresponding centerline-based second-order polynomial from the target sub-region mask. The fitted parameters are then applied directly to warp the source sub-region image to approximate the target sub-region image. To ensure spatial consistency, any warped pixels falling outside the target mask boundary are suppressed through cropping, and vacant regions within the target mask are filled via simple interpolation algorithms, such as bilinear interpolation. 
Let $C_p^i$ and $C_t^i$ be the representative curves (e.g., medial axis) extracted from $M_{\text{pr}}^i$, $M_{\text{tr}}^i$, respectively. We fit a quadratic warping function $f_i(x) = ax^2 + bx + c$ as summarized in Equation \ref{eqn:7} and Equation \ref{eqn:8} to find the parameters that enable fitting the source curve to the target curve:


\begin{equation}
    f_i(x_{p}^{j}) \approx y_{t}^{j},
    \qquad
    C_p=\{(x_p^j,y_p^j)\}_{j=1}^{N},
    \qquad
    C_t^i=\{(x_t^ij,y_t^ij)\}_{j=1}^{N}
    \label{eqn:7}
\end{equation}

\begin{equation}
    (a_p,b_p,c_p)
=
\arg\min_{a,b,c}
\sum_{j=1}^{N}
\left(
y_p^j-\left(a(x_p^j)^2+bx_p^j+c\right)
\right)^2
    \label{eqn:8}
\end{equation}

\begin{equation}
    (a_t,b_t,c_t)
=
\arg\min_{a,b,c}
\sum_{j=1}^{N}
\left(
y_t^{ij}-\left(a(x_t^{ij})^2+bx_t^{ij}+c\right)
\right)^2
    \label{eqn:9}
\end{equation}

\begin{equation}
    \hat{y}_p(x)=\mathrm{CubicSpline}(x_p^j,f_p(x_p^j))(x),
    \qquad
    \hat{y}_t^i(x)=\mathrm{CubicSpline}(x_t^{ij},f_t(x_t^{ij}))(x)
    \label{eqn:10}
\end{equation}

\begin{equation}
\Delta y^i(x)
=
\hat{y}_p(x)-\hat{y}_t^i(x)
\label{eqn:11}
\end{equation}

\begin{equation}
I_{\mathrm{warp}}^i(x_p,y_p)
=
\left(x_p,\;y_p+\Delta y^i(x)\right)
\label{eqn:13}
\end{equation}

In the Equation \ref{eqn:7}, $i$ is the index of the frames, $j$ represents the index of the divided sub-regions in a single frame, and $N$ is the total number of sub-region pairs. We fit quadratic functions \(f_s\) and \(f_t\) to these centerlines by least squares as in the Equation \ref{eqn:8} and Equation \ref{eqn:9}, and then resample the fitted curves using cubic spline interpolation \cite{66} as in the Equation \ref{eqn:10}. This procedure samples the common set of x coordinates from both source and target sub-regions' point sets, then recalculates both curves, enabling a more precise comparison. The vertical displacement \(\Delta y(x)\) is computed as the difference between the resampled source and target curves in Equation \ref{eqn:11}. Finally, the source body-part image is warped by remapping each pixel \((x,y)\) to \((x,y+\Delta y(x))\) as in the Equation \ref{eqn:13}. $I_{\mathrm{warp}}^i(x_p,y_p)$ is the warped, united sub-regions for the $i^{th}$ frame.

The second warping method is designed to handle cases where the previously mentioned geometric fitting is unreliable due to a (1) large gap between source and target sub-region pairs, (2) complicated occlusions to the sub-region, such as clothes, (3) overly fragmented masks, to the extent that it is hard to discern their corresponding body parts. Instead of estimating a parametric mapping, we adopt a stochastic region-level synthesis approach. Given a target sub-region mask $M_{\text{tr}}^i$, we randomly sample pixels from the corresponding source sub-region image $R_{\text{p}}^i$ and redistribute them into the shape of $M_{\text{tr}}^i$, preserving the local appearance while adapting to the target region geometry. This strategy enables coarse yet visually plausible alignment in cases where curve-based warping may introduce artifacts or distortions.
In practice, we apply this method selectively for sub-regions where the aforementioned curve-based warping fails, allowing robust editing across a wide range of motion variations.

\begin{table*}
\centering
\caption{Quantitative comparisons on 20 videos. L-S, L-N, L-T, L-B, and L-P indicate LPIPS-S, LPIPS-N, LPIPS-T, LPIPS-B, and LPIPS-P, respectively.}
\label{tab:quantitative}
\begin{tabular}{ccccccc}
\toprule
\textbf{Method} & \textbf{L-S} ($\downarrow$) & \textbf{L-N} ($\downarrow$) & \textbf{L-T} ($\downarrow$)  & \textbf{L-B} ($\downarrow$) & \textbf{L-P} ($\downarrow$) &  \textbf{CLIP} ($\uparrow$)\\
\midrule
Source Video & $\leq$0.001 & 0.082 & 0.708 & $\leq$0.001 & 0.264 & 28.63\\
Target Motion Prior & 0.709 & 0.053 & $\leq$0.001 & 0.659 & $\leq$0.001 & 24.52\\
\midrule
Follow-Your-Pose \cite{52} & 0.562 & 0.163 & 0.709 & 0.501 & 0.153 & 28.00\\
ControlVideo \cite{2} & 0.330 & 0.070 & 0.768 & 0.362 & 0.266 & 29.52 \\
MasaCtrl \cite{55} & 0.514 & 0.097 & 0.566 & 0.428 & 0.123 & 27.94\\
MotionDirector \cite{57} & 0.605 & 0.076 & 0.695 & 0.581 & 0.285 & 29.47\\
MotionEditor \cite{10} & 0.310 & 0.146 & 0.668  & 0.252 & 0.094 & 29.29\\
\midrule
\textbf{TeleMorpher (Ours)} & \textbf{0.289} & \textbf{0.099} & \textbf{0.655} & \textbf{$\leq$0.001} & \textbf{0.074} & \textbf{30.01}\\
\bottomrule
\end{tabular}
\end{table*}

\begin{table}
\centering
\caption{User preference ratio for TeleMorpher in comparisons with each baseline. A higher value indicates a stronger user preference for our TeleMorpher. M-A, A-A, and T-A correspond to motion alignment, appearance alignment, and textual alignment, respectively.}
\label{tab:user}
\begin{tabular}{cccc}
\toprule
\textbf{Method} & \textbf{M-A} & \textbf{A-A} & \textbf{T-A}\\
\midrule
Follow-Your-Pose \cite{52} &97.1\%&94.6\% & 91.5\%\\
ControlVideo \cite{2} & 84.2\% & 69.4\% & 83.3\%\\
MasaCtrl \cite{55} & 92.5\%& 94.5\%& 91.2\%\\
MotionDirector \cite{57} & 93.7\%& 93.3\%& 85.2\%\\
MotionEditor \cite{10} & 75.3\%& 81.2\% & 79.9\%\\
\bottomrule
\end{tabular}
\end{table}

\subsection{Protagonist-Guided Inference of Motion Editor}
\label{subsec:44}

To reduce the gap between the source and the target motions and to better preserve the protagonist's appearance during inference—particularly in cases with significant pose or motion discrepancies—we inject the protagonist condition directly into the UNet's attention layers. 

Specifically, we selectively overwrite a portion of the attention keys and values with the protagonist’s features using a lightweight blending operation. Let $K, V \in \mathbb{R}^{B \times h \times w \times C}$ be the key and value tensors, and let $P \in \mathbb{R}^{F \times h \times w \times 3}$ denote the protagonist condition. $B$ is the batch dimension of the attention block, and $h$ and $w$ are the height and width of the attention block. $C$ is the number of channels. We first repeat $P$ along the batch dimension to match $B$, resulting in $P_{\text{re}} \in \mathbb{R}^{B \times h \times w \times 3}$. Then, we overwrite a subset of feature channels in $K$ and $V$ using repeated RGB values:

\begin{equation}
    K[:, :, :, \mathcal{I}] \leftarrow \operatorname{Repeat}(P, k),
    \qquad
    V[:, :, :, \mathcal{I}] \leftarrow \operatorname{Repeat}(P, k)
    \label{eqn:14}
\end{equation}

In the Equation \ref{eqn:14}, $\text{Repeat}(P_{\text{re}}, k) \in \mathbb{R}^{B \times h \times w \times 3k}$ denotes the repeated RGB feature $k$ times, that is empirically adjusted along the channel axis. The ratio $\frac{3k}{C}$ is treated as a hyperparameter to control the guidance strength during inference. \(P\) denotes the protagonist guidance feature obtained from $I_{\mathrm{warp}}$, and \(\mathcal{I}\) denotes a randomly selected subset of feature-channel indices in the key and value tensors. We then replace the selected key and value channels with the repeated protagonist guidance feature, thereby injecting protagonist-specific appearance and motion cues into a subset of the attention representation.

To further reinforce both the target conditioning effect and appearance consistency, we further perform a lightweight \textit{mixup} between the input video $X \in \mathbb{R}^{F \times h \times w \times 3}$ and the protagonist condition $P$ to further boost the effect of the protagonist guidance. This \textit{mixup} is described in the Equation \ref{eqn:15}:

\begin{equation}
\tilde{X} = \lambda X + (1 - \lambda) P, \quad \lambda \in [0, 1]
\label{eqn:15}
\end{equation}

Here, $\lambda$ is a mixing coefficient (e.g., 0.8), which is treated as a hyperparameter.

\section{Experiments}
\subsection{Implementation Details}

Our proposed TeleMorpher is heavily based on the MotionEditor \cite{10}, which is a pioneering open-source motion editing framework. To disentangle the protagonist from the background, we leveraged Segment Anything \cite{43}. For the background inpainting, we used Inpaint Anything \cite{44} and Fotor \cite{48}. For the pose extraction, we employed OpenPose \cite{47}. Motion priors are initially collected from the text2motion \cite{49}, and then we synthesized motion priors through MotionEditor to systematically cover various cases. We evaluate our model on in-the-wild videos collected from YouTube, Hailuo AI \cite{50}, and Pexels \cite{51}. The frame resolution is 512 x 512, and the frame length was 5 due to limited resources. We set the training and inference hyperparameters following MotionEditor settings, which leverage DDIM, Stable Diffusion, and classifier-free guidance. We performed motion editing on a half-version of a single NVIDIA A100 GPU. 

\subsection{Motion Editing Results}

\begin{figure}[ht]
\begin{center}
\includegraphics[width=4.0in]{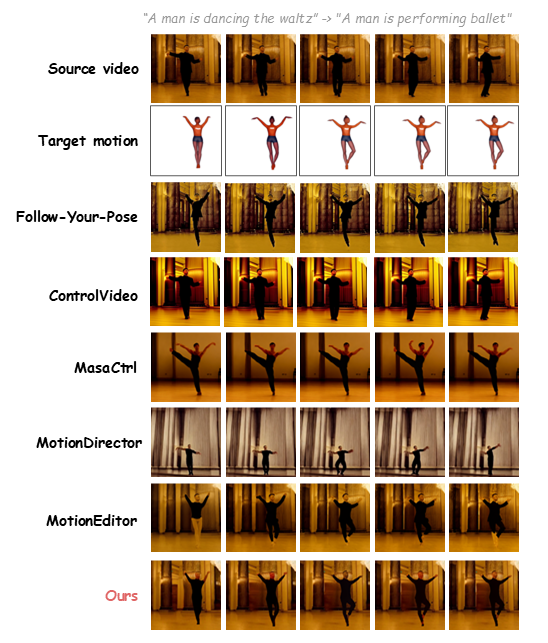}
\caption{Qualitative comparisons between TeleMorpher and recent state-of-the-art video editing approaches demonstrate that our method achieves more precise motion editing while effectively preserving the appearance.}
\label{fig:motion9}
\end{center}
\end{figure}

Figure \ref{fig:results} and Figure \ref{fig:motion9} shows the representative results of TeleMorpher's motion editing. We demonstrate the superiority of our proposed method across 20 cases collected arbitrarily, thereby showing TeleMorpher's performance compared to other baselines in the selected use cases. It includes motion, location editing, and simultaneous editing. We can see that TeleMorpher consistently edits location and motion simultaneously across diverse domains, better preserving the appearances of the protagonist and background.

\subsection{Comparison with State-of-the-Art Methods}

\textbf{Competitors.} We compare our TeleMorpher with recent state-of-the-art approaches to demonstrate its effectiveness in motion editing. The baseline methods are summarized as follows: (1) Follow-Your-Pose \cite{52}, which introduces a spatial pose adapter and a two-stage training recipe for pose-guided video generation. (2) ControlVideo \cite{2}, which incorporates fully cross-frame attention to generate videos by treating them as stitched spatial sequences. (3) MasaCtrl \cite{55} is a tuning-free image editor that can also be applied to video generation by using it for all video frames. (4) MotionDirector \cite{57} adapts dual-path Low-Rank adaptations \cite{58} architecture to decouple the learning of appearance and motion with appearance-debiased temporal loss for diverse appearances with customized motions. (5) MotionEditor \cite{10}, a two-branch framework that leverages a motion adapter to integrate the motion condition with the protagonist and background appearance in a temporally coherent manner. We set the fine-tuning step to 300 for all experiments, equal to the TeleMorpher. We followed the official environment and hyperparameter settings from the public repositories available around May 2025 for all baselines except MasaCtrl. For MasaCtrl, we slightly adjusted a conditioning-related hyperparameter to obtain stable outputs while keeping the remaining settings unchanged.

\noindent\textbf{Qualitative results.} We present qualitative comparisons in Figure \ref{fig:motion9} and Appendix \ref{appendix:exp}. The following observations can be made from the results: (1) Follow-Your-Pose and MasaCtrl fail to preserve the appearance of the source video during motion editing. Although they manage to transform the motion according to the target, the background undergoes significant changes. (2) ControlVideo preserves both the protagonist's appearance and the background well, but exhibits poor alignment with the target motion. (3) MotionDirector, which is designed to produce diverse editing outputs, shows significant deviations in both motion and appearance, resulting in less faithful edits. (4) MotionEditor, a strong baseline, demonstrates competitive performance in preserving appearance and aligning motion. However, it suffers from flickering and temporal inconsistencies across frames. In contrast, our proposed method, TeleMorpher, achieves more accurate, controllable motion editing with more easily customizable motion priors while faithfully preserving the appearance and temporal consistency of the original video. 

\noindent\textbf{Quantitative results.} We quantitatively evaluate our method against recent approaches using a suite of widely adopted metrics \cite{53,54}. In addition to LPIPS-S, LPIPS-N, LPIPS-T, and CLIP, which were leveraged from previous work, we newly defined LPIPS-B and LPIPS-P to more clearly analyze results. The evaluation metrics are defined as follows: (1) LPIPS-S (Source) computes the Learned Perceptual Image Patch Similarity (LPIPS-S) between the source and the edited video frames, that can evaluate the ability to preserve appearance in the source video; (2) LPIPS-N (Neighbor) quantifies the perceptual distance between adjacent frames in the edited video, indicating temporal consistency; (3) LPIPS-T (Target) evaluates the perceptual distance between the edited video and the target motion prior, reflecting how well the generated output follows the target motion; (4) The newly proposed LPIPS-B (Background) measures the degree to which background content from the source video is preserved in the edited video. To solely focus on the background similarity, we employ Fotor \cite{48} to remove human subjects from all frames and compute the metric on the resulting background-only regions. LPIPS-B complements LPIPS-S by isolating background regions and excluding the intentionally modified protagonist motion from the evaluation; (5) The newly introduced LPIPS-P (Pose) computes the similarity between OpenPose skeletons between the edited output and the target motion prior. For the case where OpenPose could not adequately extract poses, we manually annotated poses for those frames. By representing both videos as skeletons, LPIPS-P reduces the influence of appearance differences and focuses primarily on motion-location alignment; (6) CLIP Score assesses the semantic alignment between the edited video and the target textual description, providing an overall measure of how well the generated motion reflects the intended instruction. We used more detailed prompts for the evaluations, encompassing the protagonist's appearance and background.

The quantitative results are presented in Table \ref{tab:quantitative}. (1) Follow-Your-Pose relatively underperforms, exhibiting poor preservation of the source appearance and background, which is reflected in high LPIPS-T and LPIPS-B. They also reported inferior motion alignment with target motion, as evidenced by high LPIPS-N, LPIPS-T, and LPIPS-P and a low CLIP score. (2) ControlVideo, despite achieving a low LPIPS-N, suggesting temporal smoothness and LPIPS-S indicating successfully preserved appearance, records significantly higher LPIPS-T and LPIPS-P values, showing that it fails to align the motion with the target motion. In particular, a similar LPIPS-P with source motion and significantly low LPIPS-N indicate they generate static motions resembling the source motions. (3) MasaCtrl shows great temporal consistency (LPIPS-N), but its elevated LPIPS-P implies poor alignment with the target motion. Low LPIPS-T but high LPIPS-P indicates that the lower LPIPS-T is due to the appearance rather than the edited motion quality. (4) Because MotionDirector is not designed to generate diverse output, they recorded high LPIPS-S, LPIPS-T, LPIPS-B, and LPIPS-P. However, they achieve great temporal consistency. (5) MotionEditor also falls short across most metrics compared to our approach. However, it is noteworthy that they obtained relatively low LPIPS-S, LPIPS-B, and LPIPS-P scores, indicating decent appearance preservation and motion alignment. (6) Lastly, although TeleMorpher did not achieve state-of-the-art scores in all metrics, it achieves the best LPIPS-S, LPIPS-B, LPIPS-P, and CLIP scores. This result is expected because the proposed pipeline explicitly preserves and restores the background through disentanglement and uses the pre-trained inpainting models. These results validate the superior appearance preservation of the proposed method, the alignment of the motion with the target motion, and decent temporal coherence.

We additionally conduct a user study to assess human preferences between our method and existing competitors, selecting the better editing results, as shown in Figure \ref{fig:user}. The design of our user study closely follows the protocol used by MotionEditor. For the human evaluation, we randomly selected 10 cases from the 20 test cases. The study involved 25 participants from diverse backgrounds. For each case, participants were first shown the source video and the corresponding target motion. Subsequently, they were presented with two motion-edited videos—one generated by our proposed TeleMorpher and the other by a baseline method—in randomized order. Participants were then asked to respond to the following questions: (I) Which video exhibits better alignment with the target motion? (M-A) (II) Which video better preserves the appearance of the source video? (A-A) (III) Which video better aligns with the given text prompt? (T-A) As shown in Table \ref{tab:user}, our method consistently outperforms other approaches in terms of subjective user preference.

\subsection{Ablation Study}

To demonstrate the effectiveness of TeleMorpher's core components, we conduct an ablation study. Following the previous study \cite{10}, we qualitatively analyze the results. Figure \ref{fig:ablation} shows the results of applying each module. The first column shows that motion editing without a background achieves better results than with a background. The second column in the figure shows that pose guidance alleviated flickering and improved temporal coherence by enhancing the conditioning power of target motion. The third column also shows that empowering the target motion condition via the \textit{mixup} strategy of the input and protagonist conditions further enhanced motion fidelity and temporal coherence.

\section{Limitation}

Although TeleMorpher demonstrates better performance in motion editing, several limitations remain. First, we observed occasional color shifts of the protagonist in the edited videos. Although these changes have minimal impact on the structural fidelity of the motion, they affect the appearance consistency and quality. This limitation could be mitigated by integrating post-color correction techniques applied after motion editing. Second, due to limitations in computational resources, our experiments were conducted on short video clips. One option to mitigate this limitation is to segment the video into shorter clips and apply our method iteratively.

\section{Acknowledgement}
I would like to express my sincere gratitude to the support( \href{https://drive.google.com/file/d/1vbXr0Up6HFGR1iZFCVotKdeHXkPJfWbN/view?usp=drive_link}{\texttt{Appreciation link}}).

\section{Conclusion}

Via analyzing fundamental reasons for the degradation of the performance of the motion-location editing, this study introduces TeleMorpher, a new motion editing pipeline. This pipeline improves both controllability and visual fidelity by leveraging motion priors and protagonist guidance, that can more accurately edit the motion of a protagonist in the source video while better preserving the appearance of the source video. Unlike previous motion editing methods that rely on reference videos, leveraging the motion priors generated from highly controllable 3D avatar systems enables users to specify desired motions through virtual avatars and directly transfer them into real-world videos. Therefore, TeleMorpher bridges the controllability of modern avatar platforms, allowing the motion design flexibility and efficiency of virtual worlds to be applied to existing human videos without additional rendering, animation, or manual control. In addition, the use of skeletons and segmentation masks to generate motion-location guidance for the pose warping procedure lays the groundwork for the scalable, precise editing of various video and image structures in the future. This can include precise object editing, protagonist-object interaction editing, and multi-protagonist editing.

\bibliographystyle{unsrt}
\bibliography{references}

\pagebreak

\appendix

\section{Appendix}

\begin{figure*}
\begin{center}
\includegraphics[width=\columnwidth]{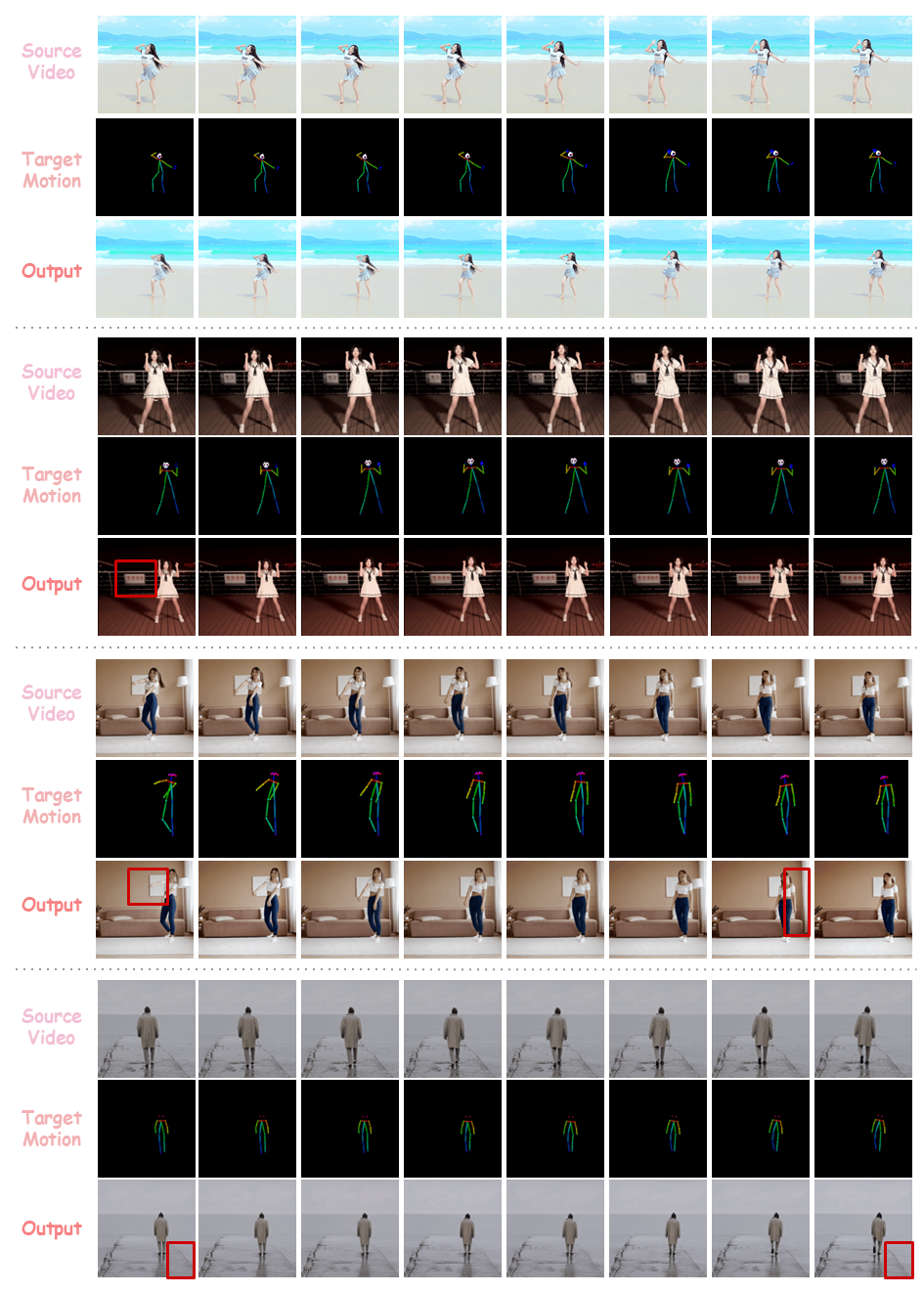}
\caption{Impact of location transformation on video editing quality. This figure illustrates the performance drop when location changes are applied to the protagonist during motion editing. Compared to the source video, edited results show degradation in appearance consistency and temporal stability, indicating that location transformation is a significant factor contributing to the difficulty of video editing. The text prompts for the four videos are "A girl is dancing" for the top three videos and "A man is walking" for the last video.}
\label{fig:location}
\end{center}
\end{figure*}

\begin{figure*}
\begin{center}
\includegraphics[width=\columnwidth]{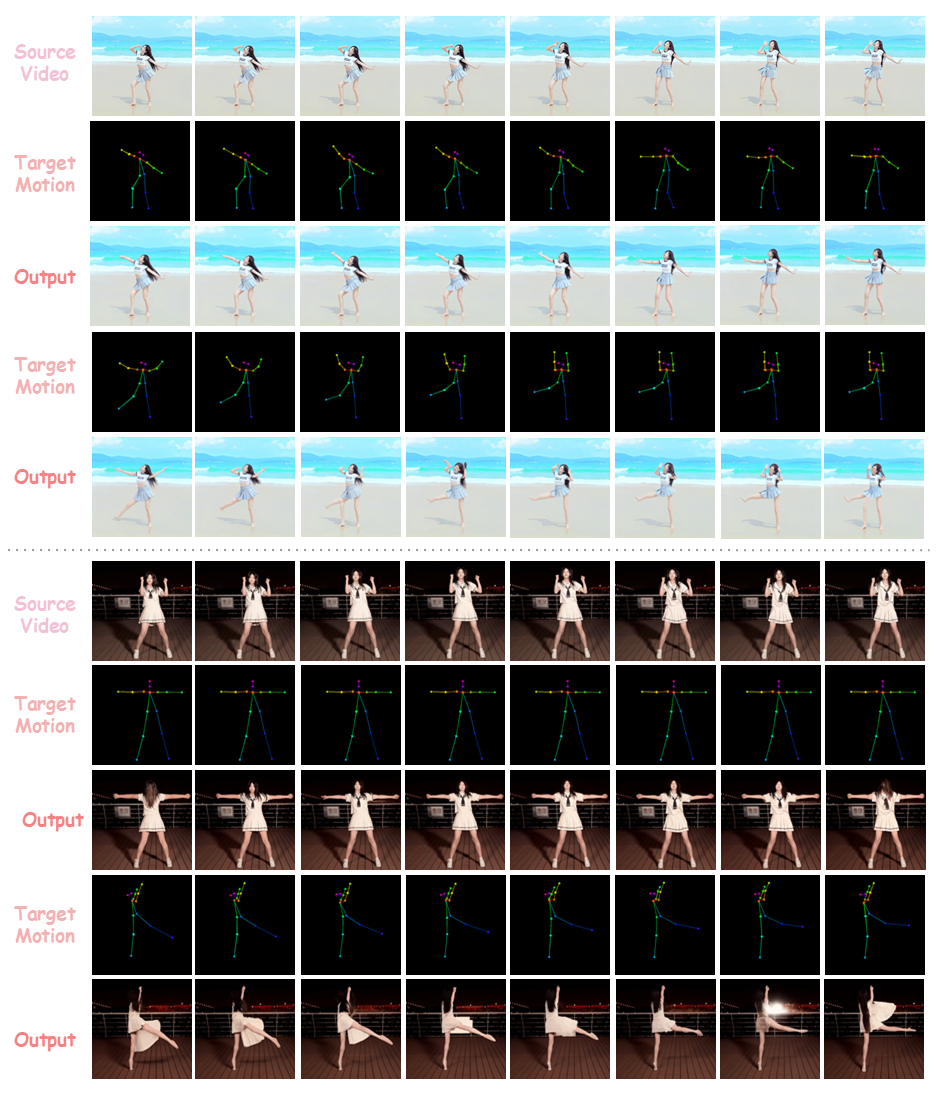}
\caption{Edited video with various motion differences. We compared relatively smaller motion differences and larger differences between the two videos. The first cases (top of each case) in the two videos only change upper-body movement (level 1), whereas the second cases (bottom of each case) change the entire body movement (level 2). We can see that when the motion gap is greater (level 2), the quality of edited videos is degraded compared to when it is smaller (level 1). The text prompt for the two videos is "A girl is dancing". }
\label{fig:motion1}
\end{center}
\end{figure*}

\begin{figure}[ht]
\begin{center}
\includegraphics[width=5in]{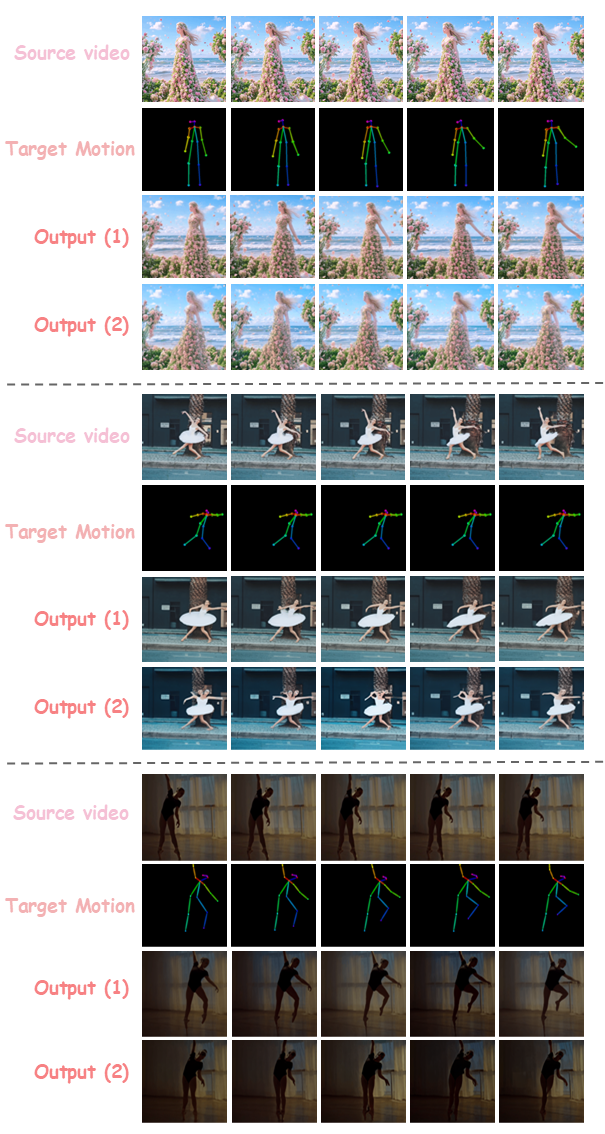}
\caption{Motion editing results with two resolutions, which are (512 x 512) and (256 x 256). The "Output (1)" is the edited video with a resolution of (512 x 512), whereas the "Output (2)" indicates the edited video with a resolution of (256 x 256). The prompts are "A girl raises her arm up", "A girl performs ballet with her arms extended straight out", "A girl is doing a ballet jump with her arms opened and her leg folded" from the top.}
\label{fig:motion1_5}
\end{center}
\end{figure}

\begin{figure*}
\begin{center}
\includegraphics[width=\columnwidth]{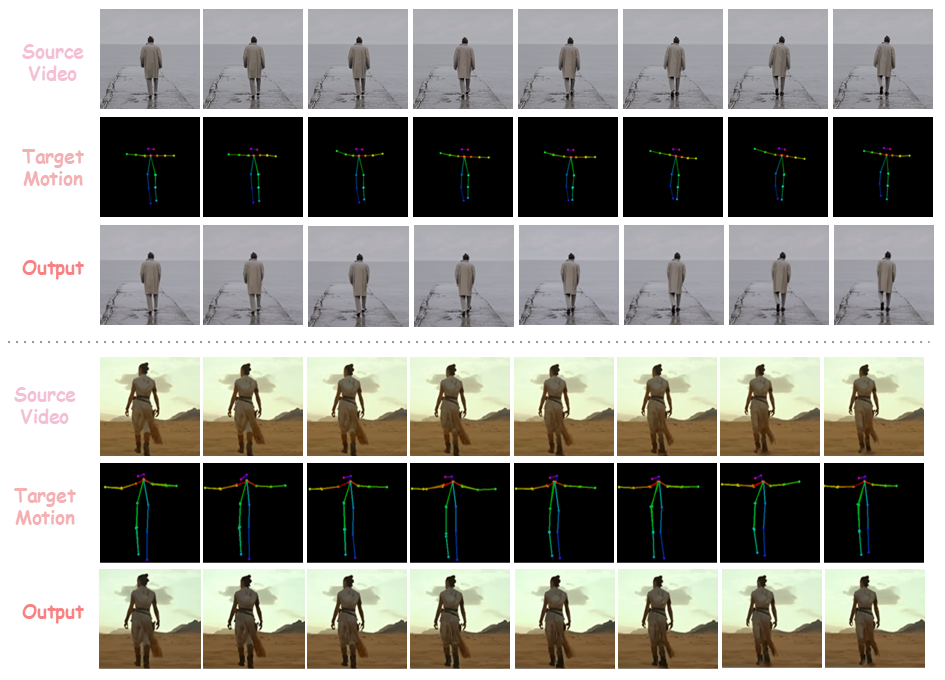}
\caption{Motion editing results with ambiguous videos, which have similar colors between the protagonists and the backgrounds. The bottom case is more ambiguous because of its inherent blur due to its low resolution (256 x 256). We can see that in both cases, the motion transformation has failed resulting in similar video with the source video. The prompts are "A man is walking" and "A woman is walking" from the top.}
\label{fig:motion2}
\end{center}
\end{figure*}

\begin{figure*}
\begin{center}
\includegraphics[width=\columnwidth]{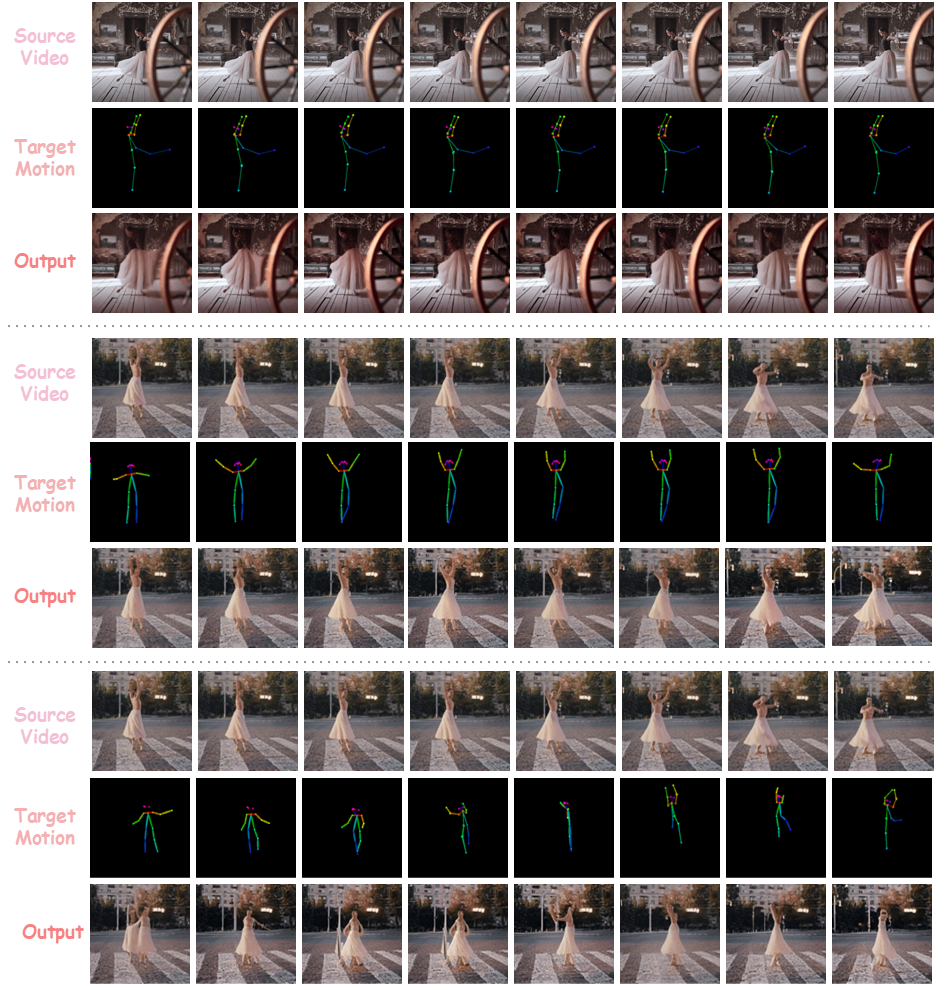}
\caption{Motion editing results under highly challenging scenarios where multiple degradation factors co-exist. The top case combines challenging background, high human-background similarity, and a large motion gap. The middle case involves relatively small motion changes but is affected by strong camera motion, low resolution (256×256). The bottom case is the most difficult, with compounded factors including a large motion gap, location shift, and motion dynamicity. In all cases, motion transformation quality significantly drops, confirming the compounded impact of multiple editing constraints. The prompt is "A girl is performing ballet".}
\label{fig:motion3}
\end{center}
\end{figure*}

\section{Problem Definition}
\label{appendix:problem}

We conducted a series of experiments to empirically verify how the factors identified in Section 3 contribute to the degradation of video editing quality. We analyze the impact of gap (location, motion), ambiguity (resolution, human-background similarity), and information amount (background complexity, motion dynamicity, camera movement) in motion editing quality, through several experiments. The results are presented in Figures \ref{fig:location} through Figure \ref{fig:motion3} and Table \ref{tab:location} through Table \ref{tab:motion3}, each corresponding to a different challenge category. 

\subsection{Gap}

\textbf{Location gap analysis.} The first experiment, analyzes the impact of location transformation. By comparing the quality of the edited video after applying location shifts to that of the source video, we demonstrate that location editing introduces significant challenges, leading to a noticeable drop in generation quality. As shown in Table \ref{tab:location}, location transformations introduce noticeable artifacts, resulting in higher LPIPS scores. This validates that controlling location editing is critical for degrading the overall quality of edited videos.

\noindent\textbf{Motion gap analysis.} For the second experiment, we evaluate the effect of motion gap by applying Level 1 and Level 2 motion transformations. All other conditions were set to Level 1 to isolate the impact of motion differences. In Level 1, we modified only the upper-body motion while keeping the lower-body motion consistent with the source video. Specifically, both source and target videos featured upper-body motion in which the arms were fully extended to the sides. In contrast, Level 2 involved significant changes in both upper and lower-body movements. As shown in Figure \ref{fig:motion1} and Table \ref{tab:motion1}, we observed that Level 1 motion transformations were relatively well handled, whereas Level 2 transformations led to noticeable degradation in video editing quality. This demonstrates that larger motion gaps make effective motion editing more difficult.

\subsection{Ambiguity}

\textbf{Resolution.} The third experiment examines the effect of resolution on motion-location editing performance. To assess this, we conducted motion editing on videos downscaled from (512 × 512) to (256 × 256) resolution. As shown in Table \ref{tab:motion1_5}, lowering the resolution led to a notable increase in both L-P and L-B scores, indicating significant degradation in motion alignment and appearance preservation. Figure \ref{fig:motion1_5} qualitatively compares the motion-location editing results from (512 × 512) to (256 × 256) resolution. Through this, we can see that the outputs from (256 x 256) videos have inferior motion and appearance drift compared to those from (512 x 512) videos. These findings suggest that limited spatial detail due to low resolution hinders the model’s ability to robustly transform the videos and maintain visual consistency, ultimately impairing motion-location editing quality.

\noindent \textbf{Human-background similarity.} The fourth experiment demonstrates that human-background similarity is one of the key factors that degrade motion editing quality. To validate this, we selected a scenario where the background was relatively simple, but the color and texture of the protagonist closely resembled the background, corresponding to a Level 2 human-background similarity setting. The target motion was set to an arm-extended pose, which previously showed positive results in the motion gap experiment. All other factors were set to Level 1 to isolate the effect of human-background similarity. As shown in Figure \ref{fig:motion2} and Table \ref{tab:motion2}, motion editing failed to produce noticeable changes, even at a relatively high resolution (512 x 512), with minimal differences between the pre- and post-editing motions. These results confirm that high human-background similarity significantly impairs the model’s ability to perform effective motion editing.

\begin{figure*}
\begin{center}
\includegraphics[width=\columnwidth]{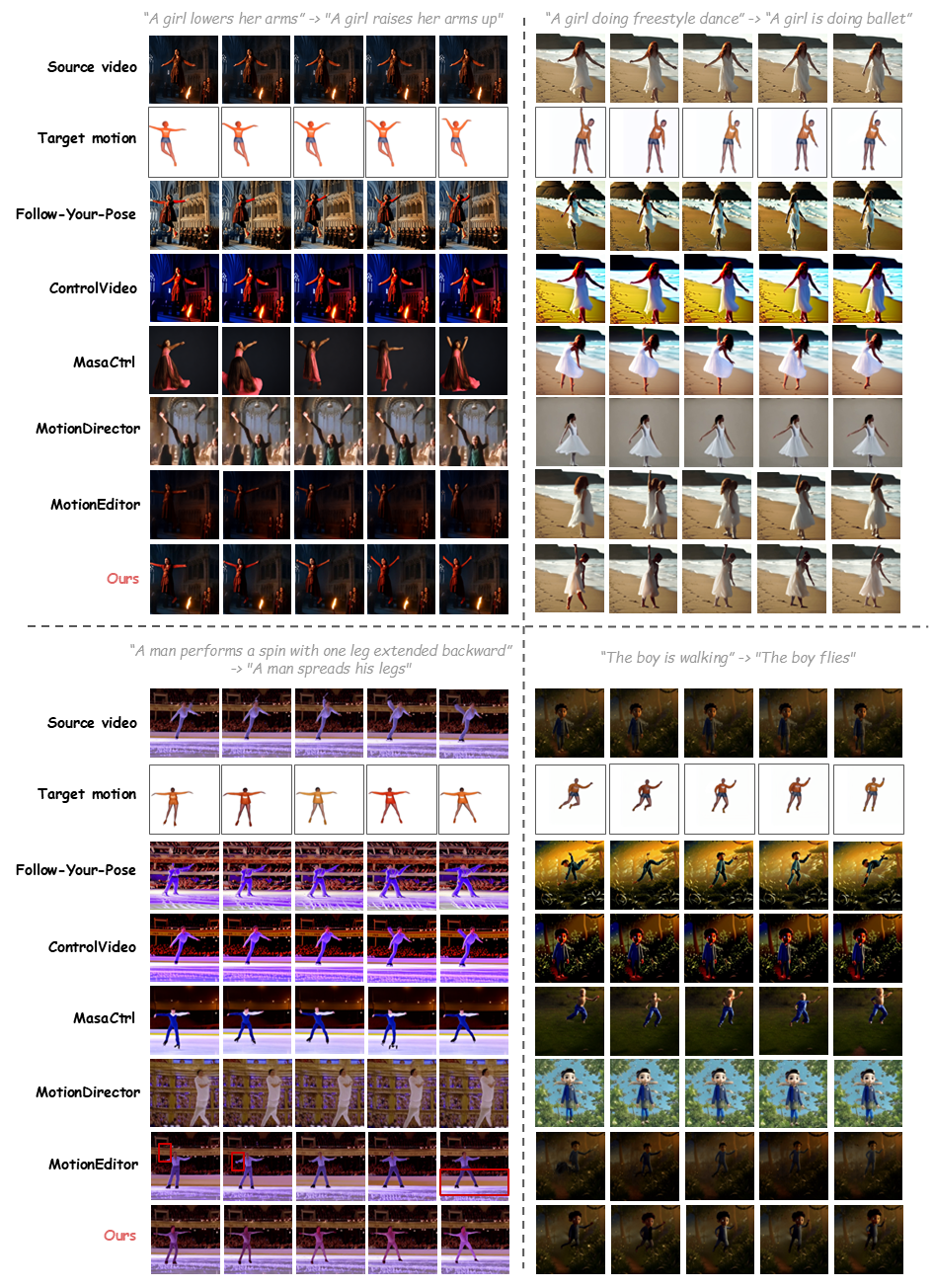}
\caption{Qualitative comparison between our proposed method, TeleMorpher, and baseline video editing models. While prior methods often suffer from inaccurate motion transformation or appearance distortion, TeleMorpher achieves more precise motion alignment and better preserves the appearance.}
\label{fig:adiitional}
\end{center}
\end{figure*}

\begin{figure*}
\begin{center}
\includegraphics[width=\columnwidth]{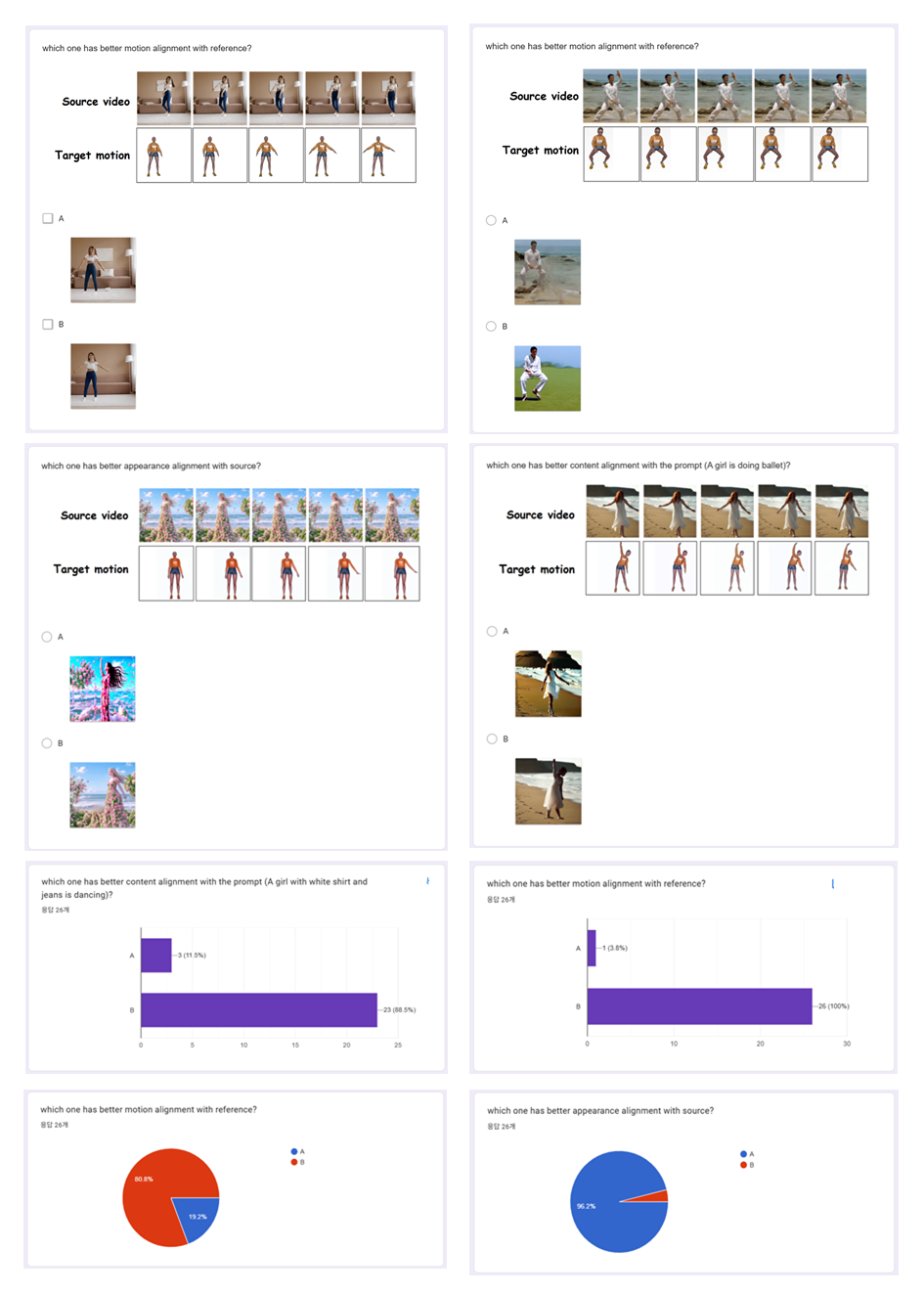}
\caption{Screenshot of the user study interface. For each case, participants were presented with the source video and a target motion. Two edited videos (ours and a baseline) were shown in randomized order. Participants answered three questions assessing motion alignment, appearance alignment, and textual alignment.}
\label{fig:user}
\end{center}
\end{figure*}

\subsection{Information Amount}

\textbf{Compound analysis.} The final experiment explores scenarios where multiple factors that degrade motion editing quality are combined, in addition to the factors mentioned earlier. In the first case of Figure \ref{fig:motion3}, the video features a challenging background (background complexity: Level 2), high human-background similarity (Level 2), and a large motion gap (Level 2), making it a highly challenging setting. The second case involves only a small motion change (motion gap: Level 1), where only the protagonist’s arm motion is edited; however, it includes significant camera movement (camera motion: Level 2) and a relatively cluttered background (background complexity: Level 2). Additionally, the video resolution is limited to 256×256, resulting in substantial visual blur (blur: Level 2). The third case builds upon the second, with added complexity from large location changes, a high motion gap, and dynamicity, categorized as Level 2.

As shown in Figure \ref{fig:motion3} and Table \ref{tab:motion3}, all three cases exhibit considerable performance degradation. Compared to the results in Tables \ref{tab:location}, \ref{tab:motion1}, and \ref{tab:motion2}, we observe a compounding effect where the combination of multiple challenging factors results in more severe deterioration. These findings further support that the proposed criteria—motion gap, human-background similarity, background complexity, and others—are critical factors that make motion editing substantially more difficult.

\subsection{Additional Experimental Results}
\label{appendix:exp}

To complement the main results presented in the paper, we provide additional qualitative comparisons between TeleMorpher and recent state-of-the-art video editing baselines. As shown in Figure \ref{fig:adiitional}, our method consistently outperforms existing approaches in terms of motion alignment and appearance preservation. Although baseline methods often suffer from distorted body parts, unnatural transitions, or inconsistent backgrounds, TeleMorpher generates more coherent, visually plausible frames. These results further demonstrate the effectiveness and robustness of our pipeline, particularly in handling challenging motion edits that involve various pose and location changes.

\begin{table*}[t]
\centering
\label{tab:folder_comparison}
\resizebox{\columnwidth}{!}{%
\begin{tabular}{l|cccc|cccccc}
\toprule
& \textbf{L-N} ↓ & \textbf{L-T} ↓ & \textbf{L-P} ↓  & \textbf{CLIP} ↑
& \textbf{L-S} ↓ & \textbf{L-N} ↓ & \textbf{L-T} ↓ & \textbf{L-B} ↓ & \textbf{L-P} ↓ & \textbf{CLIP} ↑ \\
\midrule
& \multicolumn{4}{c|}{Source Video} & \multicolumn{6}{c}{Edited Video} \\
\midrule
Case A & 0.043 & 0.628 & 0.276 & 26.92 & 0.210 & 0.051 & 0.564 & 0.033 & 0.059 & 26.00\\
Case B & 0.062 & 0.598 & 0.222 & 25.83 & 0.314 & 0.063 & 0.544 & 0.189 & 0.045 & 27.22\\
Case C & 0.061 & 0.677 & 0.272 & 26.26 & 0.300 & 0.091 & 0.606 & 0.168 & 0.035 & 26.36\\
Case D & 0.017 & 0.500 & 0.183 & 26.22 & 0.210 & 0.034 & 0.475 & 0.088 & 0.025 & 26.22\\
\bottomrule
\end{tabular}}
\caption{The video quality comparison before and after editing the location of the protagonists. The Case A-D represents the videos in Figure \ref{fig:location} from top to bottom. The text prompt of Case A-C is "A girl is dancing" and Case D is "A man is walking".}
\label{tab:location}
\end{table*}

\begin{table*}[t]
\centering
\label{tab:folder_comparison}
\resizebox{\columnwidth}{!}{%
\begin{tabular}{l|cccc|cccccc}
\toprule

 & \textbf{L-N} ↓ & \textbf{L-T} ↓ & \textbf{L-P} ↓  & \textbf{CLIP} ↑
& \textbf{L-S} ↓ & \textbf{L-N} ↓ & \textbf{L-T} ↓ & \textbf{L-B} ↓ & \textbf{L-P} ↓ & \textbf{CLIP} ↑ \\
\midrule
& \multicolumn{4}{c|}{Source Video} & \multicolumn{6}{c}{Edited Video} \\
\midrule
Case A-1 & 0.036 & 0.566 & 0.077 & 27.00 & 0.054 & 0.046 & 0.573 & 0.030 & 0.061 & 27.17\\
Case A-2 & 0.036 & 0.566 & 0.077 & 27.00 & 0.110 & 0.061 & 0.586 & 0.032 & 0.095 & 27.12\\
\midrule
Case B-1 & 0.042 & 0.571 & 0.159 & 25.85 & 0.129 & 0.050 & 0.575 & 0.107 & 0.092 & 26.36\\
Case B-2 & 0.042 & 0.571 & 0.159 & 25.85 & 0.306 & 0.100 & 0.581 & 0.177 & 0.143 & 28.22\\
\bottomrule
\end{tabular}}
\caption{The video quality comparison before and after editing the motion of the protagonists in the Figure \ref{fig:motion1}. The videos used are not ambiguous and have less information amounts. Case A represents the video with the blue background, and Case B represents the video with the girl in the white dress. 1 and 2 indicate the first and second motion editing for each video from top to bottom.}
\label{tab:motion1}
\end{table*}

\begin{table*}[t]
\centering
\label{tab:folder_comparison}
\resizebox{\columnwidth}{!}{%
\begin{tabular}{l|cccc|cccccc}
\toprule
 & \textbf{L-N} ↓ & \textbf{L-T} ↓ & \textbf{L-P} ↓  & \textbf{CLIP} ↑
& \textbf{L-S} ↓ & \textbf{L-N} ↓ & \textbf{L-T} ↓ & \textbf{L-B} ↓ & \textbf{L-P} ↓ & \textbf{CLIP} ↑ \\
\midrule
& \multicolumn{4}{c|}{Source Video} & \multicolumn{6}{c}{Edited Video} \\
\midrule
Case A-1 & 0.069 & 0.821 & 0.302 & 22.80 & 0.358 & 0.154 & 0.812 & 0.285 & $\leq$0.001 & 22.21\\
Case A-2 & 0.069 & 0.821 & 0.302 & 22.80 & 0.394 & 0.132 & 0.829 & 0.331 & 0.182 & 23.21\\
\midrule
Case B-1 & 0.158 & 0.642 & 0.189 & 33.03 & 0.222 & 0.137 & 0.605 & 0.224 & 0.060 & 33.00\\
Case B-2 & 0.158 & 0.642 & 0.189 & 33.03 & 0.311 & 0.149 & 0.617 & 0.283 & 0.194 & 33.08\\
\midrule
Case C-1 & 0.039 & 0.536 & 0.430 & 30.98 & 0.368 & 0.088 & 0.344 & 0.162 & 0.058 & 30.66 \\
Case C-2 & 0.039 & 0.536 & 0.430 & 30.98 & 0.367 & 0.086 & 0.331 & 0.172 & 0.194 & 31.29 \\
\bottomrule
\end{tabular}}
\caption{The video quality comparison with the two resolutions, which are (512 x 512) and (256 x 256), in the Figure \ref{fig:motion1_5}. Case A-C represents the videos from the top. Case A-1, Case B-1, and Case C-1 are from the (512 x 512) resolution videos, and Case A-2, Case B-2, and Case C-2 are obtained from (256 x 256) resolution videos. The prompts are "A girl raises her arm up", "A girl performs ballet with her arms extended straight out", "A girl is doing a ballet jump with her arms opened and her leg folded" from the top"}
\label{tab:motion1_5}
\end{table*}

\begin{table*}[t]
\centering
\label{tab:folder_comparison}
\resizebox{\columnwidth}{!}{%
\begin{tabular}{l|cccc|cccccc}
\toprule

& \textbf{L-N} ↓ & \textbf{L-T} ↓ & \textbf{L-P} ↓  & \textbf{CLIP} ↑
& \textbf{L-S} ↓ & \textbf{L-N} ↓ & \textbf{L-T} ↓ & \textbf{L-B} ↓ & \textbf{L-P} ↓ & \textbf{CLIP} ↑ \\
\midrule
& \multicolumn{4}{c|}{Source Video} & \multicolumn{6}{c}{Edited Video} \\
\midrule
Case A & 0.016 & 0.491 & 0.110 &26.12 &0.039 & 0.027 & 0.501 & 0.036 & 0.105 & 26.09\\
Case B & 0.029 & 0.596 & 0.225 & 24.56 & 0.045 & 0.036 & 0.593 & 0.025 & 0.216 & 24.60\\
\bottomrule
\end{tabular}}
\caption{The video quality comparison after the motion editing with the level 2 human-background similarity. Case A is the upper case and Case B is the bottom case in Figure \ref{fig:motion2}. Both Case A and Case B fail to transform motion, generating similar results with the source videos. The text prompts are "A man is walking" and "A girl is walking" from the top.}
\label{tab:motion2}
\end{table*}

\begin{table*}[t]
\centering
\label{tab:folder_comparison}
\resizebox{\columnwidth}{!}{%
\begin{tabular}{l|cccc|cccccc}
\toprule
& \textbf{L-N} ↓ & \textbf{L-T} ↓ & \textbf{L-P} ↓  & \textbf{CLIP} ↑
& \textbf{L-S} ↓ & \textbf{L-N} ↓ & \textbf{L-T} ↓ & \textbf{L-B} ↓ & \textbf{L-P} ↓ & \textbf{CLIP} ↑ \\
\midrule
& \multicolumn{4}{c|}{Source Video} & \multicolumn{6}{c}{Edited Video} \\
\midrule
Case A & 0.084 & 0.751 & 0.204 & 25.14 & 0.233 & 0.145 & 0.790 & 0.211 & 0.183 & 24.56\\
\midrule
Case B-1 & 0.129 & 0.711 & 0.168 & 29.02 & 0.280 & 0.217 & 0.725 & 0.264 & 0.194 & 27.41\\
Case B-2 & 0.129 & 0.711 & 0.168 & 29.02 & 0.224 & 0.213 & 0.738 & 0.195 & 0.148 & 27.78\\
\bottomrule
\end{tabular}}
\caption{The video quality comparison before and after editing the motion of the protagonists in the Figure \ref{fig:motion3}. The videos used have relatively large amounts of information compared to Figure \ref{fig:location}, Figure \ref{fig:motion1}, and Figure \ref{fig:motion2}. Case A is the case at the top of the Figure \ref{fig:motion3}, and Case B indicates the two cases below with the same source video. The text prompt for the CLIP score is "A girl is performing ballet".}
\label{tab:motion3}
\end{table*}

\section{AI Disclosure} 

Large language Models (LLMs) were used for idea discussion, feedback, and language polishing. The author is solely responsible for the research design, implementation, experiments, analysis, and final conclusions.

\end{document}